\documentclass[conference]{IEEEtran}
\IEEEoverridecommandlockouts
\usepackage{cite}
\usepackage{amsmath,amssymb,amsfonts}
\usepackage{algorithmic}
\usepackage{graphicx}
\usepackage{textcomp}
\usepackage{xcolor}
\usepackage{url}
\usepackage[ruled,linesnumbered]{algorithm2e}
\SetKwComment{Comment}{/* }{ */}
\usepackage[spaces,hyphens]{xurl}
\usepackage{hyperref}

\usepackage{booktabs}
\usepackage{tabularx}
\newcolumntype{Y}{>{\centering\arraybackslash}X}
\usepackage{multirow}

\newcommand{\comment}[1]{}

\def\BibTeX{{\rm B\kern-.05em{\sc i\kern-.025em b}\kern-.08em
    T\kern-.1667em\lower.7ex\hbox{E}\kern-.125emX}}
\begin{document}

\title{Federated Survival Forests
}


\author{\IEEEauthorblockN{Alberto Archetti}
\IEEEauthorblockA{\textit{DEIB}\\
\textit{Politecnico di Milano}\\
Milan, Italy\\
alberto.archetti@polito.it}
\and
\IEEEauthorblockN{Matteo Matteucci}
\IEEEauthorblockA{\textit{DEIB}\\
\textit{Politecnico di Milano}\\
Milan, Italy\\
matteo.matteucci@polimi.it}
}

\maketitle

\begin{abstract}
Survival analysis is a subfield of statistics concerned with modeling the occurrence time of a particular event of interest for a population. 
Survival analysis found widespread applications in healthcare, engineering, and social sciences. 
However, real-world applications involve survival datasets that are distributed, incomplete, censored, and confidential. 
In this context, federated learning can tremendously improve the performance of survival analysis applications. 
Federated learning provides a set of privacy-preserving techniques to jointly train machine learning models on multiple datasets without compromising user privacy, leading to a better generalization performance. 
However, despite the widespread development of federated learning in recent AI research, few studies focus on federated survival analysis. 
In this work, we present a novel federated algorithm for survival analysis based on one of the most successful survival models, the random survival forest. 
We call the proposed method Federated Survival Forest (FedSurF). 
With a single communication round, FedSurF obtains a discriminative power comparable to deep-learning-based federated models trained over hundreds of federated iterations. 
Moreover, FedSurF retains all the advantages of random forests, namely low computational cost and natural handling of missing values and incomplete datasets. 
These advantages are especially desirable in real-world federated environments with multiple small datasets stored on devices with low computational capabilities. 
Numerical experiments compare FedSurF with state-of-the-art survival models in federated networks, showing how FedSurF outperforms deep-learning-based federated algorithms in realistic environments with non-identically distributed data.
 
\end{abstract}



\section{Introduction}
\label{sec:introduction}

Survival analysis~\cite{klein2003survival,wang2019machine}, also known as time-to-event analysis, is a subfield of statistics concerned with modeling the time until an event of interest occurs for a population of individuals. 
In particular, a survival model builds on statistical and machine learning techniques to predict a survival function $S(t)$.
This function evaluates the probability of a subject not experiencing the event up to time $t$. 
Survival analysis found widespread success in healthcare, engineering, economics, and social science applications~\cite{wang2019machine}.
However, in real-world scenarios survival data are often distributed, inaccurate, and incomplete~\cite{andreux2020federated,rieke2020future}.
In addition, survival data may contain considerable proportions of censored samples, i.e., data records for which an individual has not experienced the event yet.
Even if survival models are suited to extract all the useful information from censored samples, high censorship percentages in datasets still pose severe challenges for survival models to succeed.
One of the possible solutions is to increase the number of data samples available for training.
However, most survival applications rely on confidential data, which is private and non-sharable due to security or regulatory constraints~\cite{andreux2020federated,rieke2020future}.

In this context, Federated Learning (FL)~\cite{li2020federated,kairouz2021advances} can significantly improve the success of survival applications in real-world scenarios.
FL is a machine learning field in which a set of clients, each holding a private dataset, collaborate to train a machine learning model under the coordination of a central server. 
The crucial difference between distributed learning and FL is that private data information never leaves the device in which it is collected and stored.
FL operates by iteratively exchanging model parameters between the clients and the central server to build an average model with good performance for all the clients in the federation.
In this way, FL allows multiple parties to collectively train a machine learning model without leaking private data information stored on local devices by design. 
Federated models exhibit better generalization performance than their local counterparts since they can train on a large and representative data pool.

FL can support survival analysis in overcoming the limitations of scarce, censored, and confidential survival data. 
To this end, the field of federated survival analysis investigates the techniques to integrate survival models into federated algorithms. 
Among the works related to federated survival analysis, most focus on Cox models~\cite{lu2015webdisco,andreux2020federated,learning2020duan,dssurvival2022banerjee,verticox2022dai,larynx2022hansen,accurate2022kamphorst,federated2022masciocchi,wang2022survmaximin,imakura2023dccox,zhang2023federated}.
Despite being prominent in classical survival analysis, the Cox model is based on the proportionality assumption, which may not hold in large-scale federated datasets.
Other works extend federated survival analysis to deep neural models~\cite{federated2022lu,rahimian2022practical,rahman2022fedpseudo}.
While being extremely powerful, training these models requires several communication rounds, which may hinder convergence speed and bandwidth usage.
To the best of our knowledge, ensemble learning for federated survival analysis is yet unexplored.

In this work, we propose a novel federated adaptation of one of the most successful survival models from the machine learning literature, the Random Survival Forest (RSF)~\cite{ishwaran2008random}.
We call the proposed algorithm Federated Survival Forest (FedSurF). 
In FedSurF, each client trains a survival forest on their data locally and sends a carefully chosen subset of trees composing the local forest to the central server. 
Then, the central server builds the ensemble of trees trained on the client devices.
For each client, tree selection can occur randomly with uniform probability or according to local evaluation.
Specifically, we discuss a tree sampling technique based on the Integrated Brier Score (IBS)~\cite{graf1999assessment}.

With FedSurF, our goal is to bring RSFs and their advantages with respect to the state-of-the-art survival models to the federated setting.
In particular, RSFs have a lower computational demand than deep learning models and require less hyperparameter tuning, making them less inclined to overfitting. 
Secondly, RSFs naturally deal with missing values and categorical variables, making them able to extract most of the useful information from incomplete, inaccurate, and censored local datasets. 
Thirdly, RSFs can be interpreted more naturally due to their tree-based nature.
Finally, concerning the federated training aspects, FedSurF requires a single communication round between the clients and the central server to build the final model. 
This quality makes FedSurF much more efficient than the traditional iterative algorithms to train federated deep learning models from an inter-node communication perspective.

FedSurF has been tested in federated scenarios alongside several state-of-the-art survival models.
Numerical experiments demonstrate the efficacy of the proposed algorithm to solve survival problems with a high generalization power.
Moreover, experiments in federations with non-identically distributed data show the resilience of FedSurF to realistic scenarios in which each client exhibits a different data distribution.
In particular, FedSurF outperforms deep-learning models in most of these scenarios.

The rest of the paper is organized as follows. 
Section~\ref{sec:related-work} provides a comprehensive overview of the background and current state of the art in federated learning and survival analysis.
Section~\ref{sec:federated-survival-forests} describes our proposed algorithm, FedSurF.
Section~\ref{sec:experiments} presents a thorough analysis of the experimental evidence supporting the efficacy of FedSurF in comparison to existing approaches. 
All experimental procedures are extensively described and the source code is made publicly available to promote reproducibility.
Finally, Section~\ref{sec:conclusion} summarizes the work.

\section{Background and Related Works}
\label{sec:related-work}

This section presents an overview of the relevant literature on survival analysis and federated learning, with a focus on the current state of research in the field of federated survival analysis.

\subsection{Survival Analysis}
\label{sec:survival-analysis}

Survival analysis, or time-to-event analysis, is a subfield of statistics that models the occurrence time of an event of interest for a population. 
The main goal of a survival problem is to estimate the event occurrence probability as a function of time.
More formally, the output of a survival model is the survival function
\begin{equation*}
    S(t | \mathbf{x}) = P(T > t | \mathbf{x})
\end{equation*}
which evaluates the probability of a particular subject not having experienced the event up to time $t$. 
The subject is characterized by a feature vector $\mathbf{x} \in \mathbb{R}^d$. 
The survival function is estimated from data using machine learning and statistical techniques starting from a survival dataset.
A survival dataset $D$ is a set of triplets $(\mathbf{x}_i,\delta_i,t_i)$, $i=1, \dots, N$.
$\mathbf{x}_i$ is the $d$-dimensional feature vector characterizing the $i$-th sample.
$\delta_i$ is the event occurrence indicator.
If $\delta_i = 1$, the subject experienced the event at time $t_i$.
Conversely, if $\delta_i = 0$, the subject did not experience the event and the sample is censored.
$t_i = \min\left\{t_i^c,t_i^e\right\}$ is the minimum between the censor time $t_i^c$ and the actual event occurrence time $t_i^e$ for the $i$-th subject.

There are three types of survival models: non-parametric, semi-parametric, and parametric.
Non-parametric models make no assumption about the underlying distribution of event times.
These models are mostly used for data visualization, as they encode the summary statistics of survival data.
Non-parametric survival models are Kaplan-Meier~\cite{kaplan1958nonparametric}, Nelson-Aalen~\cite{nelson1972theory,aalen1978nonparametric}, and Life-Table~\cite{cutler1958maximum}.

Semi-parametric models focus on modeling the instantaneous risk of experiencing the event over time.
Instead of the survival function, these models predict the hazard function
\begin{equation*}
h(t | \mathbf{x}) = \lim_{\Delta t \to 0} \frac{P(t \leq T < t + \Delta t | T\geq t, \mathbf{x})}{\Delta t}
\end{equation*}
or the cumulative hazard function
$H(t | \mathbf{x}) = \int_{0}^{t} h(\tau | \mathbf{x}) \,d\tau$.
Nevertheless, the survival function of semi-parametric models can still be obtained as
$S(t | \mathbf{x}) = e^{-H(t | \mathbf{x})}$.
The hazard function helps semi-parametric models to decouple the time-varying risk $h_0(t)$ associated with the entire population, called baseline hazard, and the time-invariant risk $h(\mathbf{x})$ related to each subject, called risk function.
The most notable example of a semi-parametric survival model is Cox Proportional Hazard (CoxPH)~\cite{cox1972regression}.
CoxPH estimates the hazard function as
\begin{equation*}
h(t | \mathbf{x}) = h_0(t)e^{\langle \beta,\mathbf{x} \rangle},
\end{equation*}
where $h_0(t)$ is the baseline hazard and the risk function is $h(\mathbf{x}) = e^{\langle \beta,\mathbf{x} \rangle}$.
$h_0(t)$ can be evaluated with the Nelson-Aalen estimator~\cite{nelson1972theory,aalen1978nonparametric}.
The CoxPH model is based on the proportionality assumption, which states that the hazard ratio of two subjects is constant over time.
DeepSurv~\cite{katzman2018deepsurv} extends the CoxPH model allowing for non-linear dependencies between the input features and the output hazard.
In DeepSurv, the risk function is $h(\mathbf{x}) = \phi_w(\mathbf{x})$, where $\phi_w$ is a single-output neural network with parameters $w$.
The parameters of semi-parametric models can be optimized using the partial log-likelihood loss function, which can be evaluated using risk ratios only.

Parametric models assume that the event occurrence follows a parametric distribution.
These models are often more expressive than semi-parametric models, as they are not limited by the proportional hazard assumption, which may not hold in most real-world scenarios.
DeepHit~\cite{lee2018deephit} is a parametric model based on time discretization which models survival times directly using neural networks with sigmoid activations.
N-MTLR~\cite{fotso2018deep} extends the Multi-Task Logistic Regression (MTLR) survival model~\cite{yu2011learning} with a non-linear dependency between the input features and the predicted probabilities.
Nnet-Survival~\cite{gensheimer2019scalable,kvamme2021continuous}, also known as Logistic Hazard, parametrizes discrete hazard functions with neural networks.
PC-Hazard~\cite{kvamme2021continuous,bender2021general} is based on a piecewise constant hazard function, so that the resulting survival function is a continuous piecewise exponential.
PC-Hazard casts survival problems as Poisson regression problems, making regression models able to solve survival tasks.
Random Survival Forest (RSF)~\cite{ishwaran2008random} recursively builds a set of binary survival trees to estimate the cumulative hazard function.
As in~\cite{breiman2017classification}, binary trees are created with bootstrapping.
At each node, the feature maximizing the hazard difference between child nodes is selected among a set of $d' \leq d$ features.
The ensemble hazard function is the average of the hazard functions in the leaf nodes of the trees.

Table~\ref{tab:survival-models} summarizes the characteristics of the models described. Their comparison in federated settings is extensively discussed in Section~\ref{sec:experiments}.
\begin{table}[t]
    \centering
    \caption{Machine learning models for survival analysis.}
    \label{tab:survival-models}
    
    \begin{tabularx}{\linewidth}{@{} X Y Y Y Y @{}}
        \toprule
        \textbf{Model} & \textbf{Linear} & \textbf{Proportional hazard} & \textbf{Differentiable} & \textbf{Continuous} \\
        \midrule
        CoxPH~\cite{cox1972regression} & $\checkmark$ & $\checkmark$ & Partially & $\checkmark$\\
        DeepSurv~\cite{katzman2018deepsurv} & $\times$ & $\checkmark$ & Partially & $\checkmark$\\
        DeepHit~\cite{lee2018deephit} & $\times$ & $\times$ & $\checkmark$ & $\times$\\
        \mbox{N-MTLR}~\cite{fotso2018deep} & $\times$ & $\times$ & $\checkmark$ & $\times$\\
        \mbox{Nnet-Survival}~\cite{gensheimer2019scalable} & $\times$ & $\times$ & $\checkmark$ & $\times$\\
        \mbox{PC-Hazard}~\cite{kvamme2021continuous} & $\times$ & $\times$ & $\checkmark$ & $\checkmark$\\
        RSF~\cite{ishwaran2008random} & $\times$ & $\times$ & $\times$ & $\checkmark$\\
        \bottomrule
    \end{tabularx}
\end{table}

\subsection{Federated Learning}
\label{sec:federated-learning}

Federated Learning (FL)~\cite{li2020federated,kairouz2021advances} is a machine learning scenario in which a set of clients collaborate to train a model under the coordination of a central server.
In FL, data never leave the device in which they are collected.
A typical federated scenario is composed of $K$ clients, each holding a private dataset $D_k, k = 1, \dots, K$.
Given a machine learning model with parameters $w$, the goal a FL algorithm is to minimize a global loss function $\mathcal{L}$ with respect to $w$, i.e.,
\begin{equation*}
    \min_w \mathcal{L}(w) = \min_w\sum\limits_{k=1}^K \lambda_k \mathcal{L}_k(w).
\end{equation*}
Here, $\mathcal{L}_k$ is the loss function computed by client $k$ using their private data $D_k$. 
$\lambda_k$ is a set of parameters that weigh the contribution of each client to the global loss.
Usually, the contribution is proportional to the number of samples $|D_k|$ in order to promote low-variance losses evaluated on more data.

The first algorithm proposed in the federated learning literature to optimize $\mathcal{L}(w)$ is Federated Averaging (FedAvg)~\cite{mcmahan2017communication}. 
FedAvg alternates a broadcast iteration and an aggregation iteration until convergence.
During the broadcast iteration, the server selects subset of $K' \leq K$ clients and sends the current model parameters $w$.
Then, these clients optimize the model on local data $D_k$ and send the updated parameters back to the server.
Usually, a small number of epochs suffices for local training.
Finally, the server updates the global model parameters by averaging the updates received from the clients. 

FedAvg performs consistently well in simulated environments, but real-world scenarios pose several challenges that hinder its generalization power and convergence speed.
In particular, real-world scenarios present heterogeneous characteristics, both in terms of computational power and data distribution~\cite{li2020federated}.
To this end, several works extend FedAvg to deal with federations presenting non-reliable communication channels and non-identically distributed data among clients~\cite{reddi2020adaptive,wang2021field,li2020fedprox,karimireddy2020scaffold,acar2021federated,caldarola2022improving}.

To test the efficacy of federated algorithms in heterogeneous federations, common benchmarking practices have been developed.
The most widely spread heterogeneous dataset collection for FL is LEAF~\cite{caldas2018leaf}.
This library includes several heterogeneous datasets for standard machine learning tasks, such as image classification and next-character prediction.
SGDE~\cite{lomurno2022sgde} provides a framework to build synthetic datasets from privacy-preserving data generators trained on the clients of the federation.
With SGDE, each generator embodies the characteristic features of each client, producing inherently heterogeneous datasets.
Finally, other works~\cite{hsu2019measuring,li2022federated} study data splitting techniques based on the Dirichlet distribution.
These techniques assign data samples from a non-federated classification dataset to each client of a federation with a controllable level of heterogeneity.

\subsection{Federated Survival Analysis}
\label{sec:federated-survival-analysis}

Federated survival analysis refers to the integration of survival models into federated algorithms. 
Most works~\cite{lu2015webdisco,andreux2020federated,learning2020duan,dssurvival2022banerjee,verticox2022dai,larynx2022hansen,accurate2022kamphorst,federated2022masciocchi,wang2022survmaximin,imakura2023dccox,zhang2023federated} focus on the federated adaptation of the CoxPH model.
In particular, the Cox partial log-likelihood loss function requires access to the entire dataset in order to be evaluated.
This \emph{non-separability} poses a significant challenge in federated applications.
Alternative formulations to the standard partial log-likelihood have been proposed in some works~\cite{andreux2020federated,zhang2023federated}, relying on survival model discretization from~\cite{kvamme2021continuous}. 
Others have explored one-shot evaluation of a surrogate likelihood~\cite{learning2020duan} or distributed adaptation of the Newton-Raphson optimization method~\cite{lu2015webdisco,verticox2022dai}.

Moving beyond Cox models, in~\cite{truly2021froelicher}, the authors propose a secure Kaplan-Meier estimation procedure for federated genomics analysis. 
Other works~\cite{federated2022lu,rahimian2022practical,rahman2022fedpseudo} focus on general parametric models, including deep neural networks. 
In~\cite{rahman2022fedpseudo}, for example, the survival problem is formulated as a regression problem using pseudo-values as target labels. 
In~\cite{federated2022lu}, on the other hand, discrete survival rates are evaluated with weakly-supervised attention modules.

Privacy is a crucial aspect of distributed healthcare applications, and several works adopt differential privacy~\cite{dwork2008differential} to protect survival models against inference attacks~\cite{federated2022lu,rahimian2022practical}. 
In particular, the authors of~\cite{rahimian2022practical} tackle the tradeoff between privacy level and model performance by adding a post-processing step to regulate the parameter update and improve the convergence of the differentially-private model. 
Other works~\cite{accurate2022kamphorst,securefedyj2022marchand} rely on secure multiparty computation to prevent data leakage in a distributed network, while homomorphic encryption~\cite{truly2021froelicher} and bootstrapping with dimensionality reduction~\cite{imakura2023dccox} are less commonly used approaches for privacy preservation.

Regarding the target application, several works focus on genomics analysis for cancer investigation~\cite{andreux2020federated,federated2022lu,terrail2022flamby}, relying on the Cancer Genome Atlas (TCGA) project. FLamby~\cite{terrail2022flamby} provides a suite of federated datasets for healthcare settings, including a survival dataset for breast cancer analysis called Fed-TCGA-BRCA. The Surveillance Epidemiology and End Results (SEER) database has also been used in several works~\cite{verticox2022dai,federated2022masciocchi}. Other studies examine stroke detection~\cite{learning2020duan}, larynx cancer~\cite{larynx2022hansen}, and COVID-19 survival rates~\cite{wang2022survmaximin}. Finally, concerning performance evaluation and result comparability in federated survival settings,~\cite{archetti2023heterogeneous} provides several techniques to split existing survival datasets among the clients of a federation with a controllable level of heterogeneity.

To the best of our knowledge, FedSurF is the first ensemble learning technique for federated survival analysis, providing a parametric model that does not rely on the proportionality assumption while requiring only a single communication round to terminate.

\section{Federated Survival Forests}
\label{sec:federated-survival-forests}

In this section, we present the proposed method for adapting the Random Survival Forest (RSF) algorithm to the federated learning setting, referred to as Federated Survival Forest (FedSurF). 
Our approach is inspired by previous works in federated ensemble learning~\cite{hauschild2022federated,gencturk2022bofrf}, which involve building an ensemble model on the central server by merging base models from local ensembles on each client. 
Specifically, FedSurF involves building a RSF on the central server by aggregating the top-performing trees from local RSF models on each client. 
To this end, two techniques for tree selection are evaluated: a uniform probability selection method and a metric-based method for promoting the best-performing trees.

The FedSurF algorithm is structured in three stages: local training, tree assignment, and tree sampling.
During the local training stage, each client $k$ independently runs the RSF algorithm using their local dataset $D_k$, resulting in a local model $M_k$. 
Each $M_k$ comprises an ensemble of survival trees $\left\{T_1, \dots, T_{N_k}\right\}$. 
To optimize the performance of $M_k$ specifically for $D_k$, the hyperparameters of each $M_k$ can be fine-tuned through cross-validation using a train-validation split on the local dataset $D_k$. 
This allows for the selection of optimal values for the number of local trees $N_k$ and the tree-building parameters (maximum tree depth, minimum samples per split, minimum samples per leaf, and maximum number of leaves).

Once each client has trained a local RSF model, the tree assignment stage involves the central server determining the number of trees that each client must transmit in order to achieve the desired total of $N_S$ trees. 
To facilitate this process, the server must be notified of the number of trees $N_k$ in each local model $M_k$. 
This message has a negligible impact on the communication channel with respect to messages containing model parameters, as it contains a single integer.
Upon receiving the number of available trees $N_k$ on each client, the server performs $N_S$ iterations incrementing a tree counter $N_k' \leq N_k$ for a randomly selected client with probability proportional to the dataset cardinality $|D_k|$. 
This non-uniform probability promotes the inclusion of trees from clients with larger datasets, similar to the weighted contribution of FedAvg~\cite{mcmahan2017communication}.
After the tree assignment stage, each client is required to send $N_k'$ trees to the server, such that the total number of trees on the server is equal to $N_S$.

In the final stage of FedSurF, clients are required to select $N_k'$ trees from their local models $M_k$ to send to the central server. 
This selection process can be performed using either a uniform sampling strategy or a strategy that weights the selection probability of each tree based on a validation metric.
To this end, we propose the use of the inverse of the Integrated Brier Score (IBS) as a validation metric. 
The IBS is a widely-used measure of the accuracy of survival models and is further discussed in Section \ref{sec:eval}.
Clients can calculate the IBS score of each local tree, denoted as $\text{\emph{IBS}}_j$, where $j = 1, \dots, N_k$, and select $N_k'$ trees with a probability proportional to $1 / \text{\emph{IBS}}_j$. 
This method is referred to as FedSurF-IBS. 
If clients instead choose to use a uniform sampling strategy, the method is simply referred to as FedSurF.
Finally, the server constructs the ensemble model $M_S$ by aggregating the trees received from each client $k$.

\begin{algorithm}[t]
\caption{Federated Survival Forest (FedSurF)}
\label{alg:fedsurf}
\begin{algorithmic}[1]
\raggedright
\STATE \textbf{Input:} $K$ clients, $D_k$, $N_S$, sampling strategy
\STATE \textbf{Output:} Ensemble of survival trees $M_S$
\STATE \textbf{Initialize (server):} $N_k' \gets 0$
\STATE \textbf{Initialize (client $k$):} $M_k \gets \emptyset, M_k' \gets \emptyset$
\FOR{$k=1$ to $K$ in parallel}
    \STATE $M_k \gets \text{RandomSurvivalForest}(D_k)$
    \STATE $N_k \gets |M_k|$
\ENDFOR
\FOR{$i=1$ to $N_S$}
    \STATE $k \gets \text{Sample}(1, \{n\}_{n=1}^K, \{|D_k|\}_{k=1}^K, \text{True})$
    \STATE $N_k' \gets N_k' + 1$
\ENDFOR
\FOR{$k=1$ to $K$ in parallel}
    \IF {sampling strategy requires IBS}
        \STATE $M_k' \gets \text{Sample}(N_k',M_k,\{ 1/\text{IBS}_j \}_{j=1}^{N_k},\text{False})$
    \ELSE
        \STATE $M_k'\gets \text{Sample}(N_k',M_k,\{ 1/N_k \}_{j=1}^{N_k}, \text{False})$
    \ENDIF
    \STATE Send $M_k'$ to the server
\ENDFOR
\STATE $M_S \gets \bigcup_{k=1}^K M_{k}'$
\RETURN $M_S$
\end{algorithmic}
\end{algorithm}

The pseudocode for FedSurF is presented in Algorithm~\ref{alg:fedsurf}. 
The function Sample($N, S, P, R$) is utilized to extract $N$ elements from the set $S$, with each element being selected with probability proportional to $P$. 
If the value of $R$ is set to True, sampling is performed with replacement.



\section{Experiments}
\label{sec:experiments}

This section presents a comprehensive set of experiments to empirically demonstrate the efficacy of FedSurF in solving survival problems with a high generalization power and resilience to non-identically distributed data.
The source code is available at 
\url{https://github.com/archettialberto/federated_survival_forests}.

\subsection{Datasets}

In this study, we conduct experiments on five survival datasets: the German Breast Cancer Study Group 2 (GBSG2)~\cite{schumacher1994randomized}, the Molecular Taxonomy of Breast Cancer International Consortium (METABRIC)~\cite{katzman2018deepsurv}, the Australian AIDS survival dataset (AIDS)~\cite{ripley2023mass}, the assay of serum-free light chain dataset (FLCHAIN)~\cite{therneau2023survival}, and the Study to Understand Prognoses Preferences Outcomes and Risks of Treatment (SUPPORT)~\cite{vanderbilt2022vanderbilt}. 
The relevant summary statistics of these datasets are presented in Table~\ref{tab:survival-datasets}.

\begin{table}[t]
    \centering
    \caption{Survival datasets included in the experiments.}
    \label{tab:survival-datasets}
    
    \begin{tabularx}{\linewidth}{@{} X Y Y Y Y @{}}
        \toprule
        \textbf{Dataset} & \textbf{Samples} & \textbf{Censored} & \textbf{Numerical features} & \textbf{Categorical features} \\
        \midrule
        GBSG2~\cite{schumacher1994randomized} & 686 & 44\% & 5 & 3\\
        METABRIC~\cite{katzman2018deepsurv} & 1904 & 58\% & 5 & 3 \\
        AIDS~\cite{ripley2023mass} & 2839 & 62\% & 1 & 3 \\
        FLCHAIN~\cite{therneau2023survival} & 7874 & 28\% & 6 & 4 \\
        SUPPORT~\cite{vanderbilt2022vanderbilt} & 9105 & 68\% & 24 & 11\\
        \bottomrule
    \end{tabularx}
\end{table}

\subsection{Techniques for Federated Simulation}
\begin{figure*}[t]
    \centering
    \includegraphics[width=\linewidth]{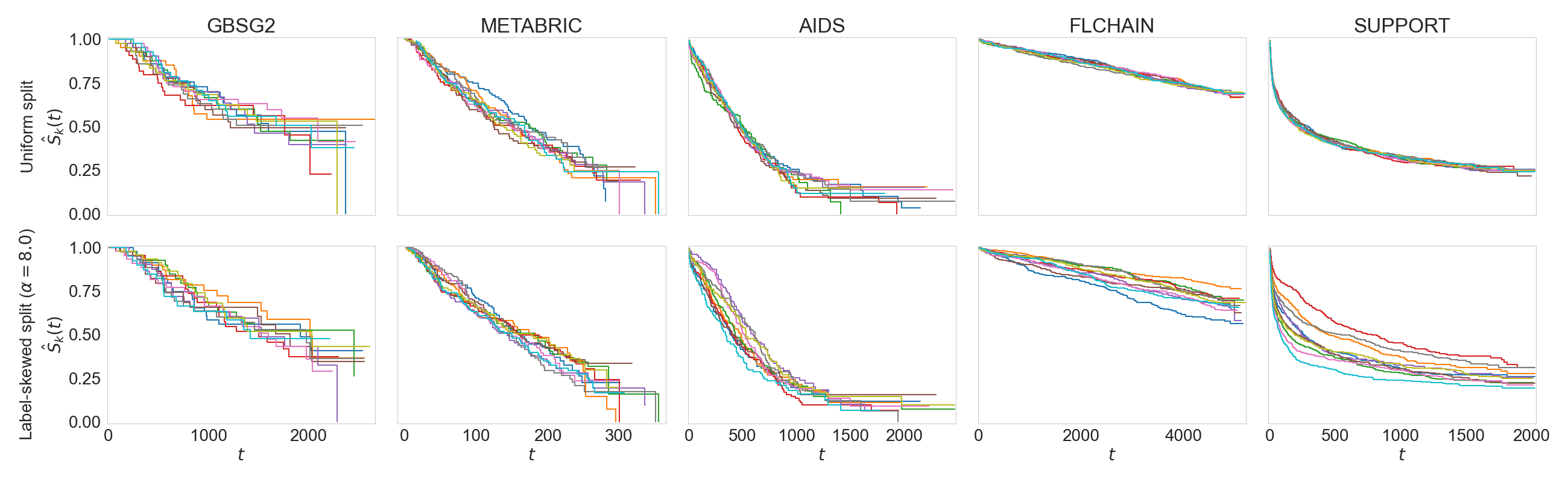}
    \caption{Kaplan-Meier estimators $\hat{S}_k(t)$ for each client dataset $D_k$ under uniform data assignment (first row) and label-skewed assignment~\cite{archetti2023heterogeneous} with $\alpha = 8.0$ (second row). Each line corresponds to a single client $k$.}
    \label{fig:km}
\end{figure*}

In the course of our experiments, we begin by partitioning a non-federated survival dataset $D$ into a training dataset and a test dataset, where the former constitutes 80\% of the total samples and the latter constitutes 20\% of the total samples. 
The test dataset is not further modified and is used exclusively for the final evaluation of each model.
Subsequently, we assign each sample in the training dataset to one of the clients in a federation.
Each federation is composed of 10 clients ($K = 10$).
To determine the client to which each sample is assigned, we employ two techniques. 
The first technique assigns each sample in the training dataset to one of the available clients with equal probability. 
This technique results in uniform data distributions among the federation clients, facilitating convergence for federated algorithms. 
The first row of Figure~\ref{fig:km} illustrates the Kaplan-Meier estimators $\hat{S}_k(t)$ of the client data under the uniform assignment technique.

The second splitting technique follows the label-skewed splitting algorithm described in~\cite{archetti2023heterogeneous}.
In this case, data samples are assigned to the federation clients to make the resulting label distributions heterogeneous.
This technique is based on the Dirichlet distribution, inspired by similar practices employed in federated datasets for classification~\cite{hsu2019measuring,li2022federated}.
The goal is to simulate realistic federated environments in which data distributions are non-uniform.
To this end, we apply the label-skewed splitting algorithm from~\cite{archetti2023heterogeneous} with $K = 10$ and $\alpha = 8.0$, imposing a minimum number of 25 samples per dataset. 
The second row of Figure~\ref{fig:km} shows the Kaplan-Meier estimators $\hat{S}_k(t)$ for each client dataset $D_k$ under label-skewed splitting.

On top of the data assignment procedures, an 80\%-20\% partition is applied to each client dataset in order to extract a local validation split for parameter tuning and early stopping.
Finally, each client is assumed to be permanently available and the communication channel is assumed to be fully reliable.

\subsection{Baseline Models}

In the experimental evaluation, we test the performance of the proposed FedSurF and FedSurF-IBS algorithms with respect to the state-of-the-art survival models from Section~\ref{sec:survival-analysis}: Cox Proportional Hazard (CoxPH)~\cite{cox1972regression}, DeepSurv~\cite{katzman2018deepsurv}, DeepHit~\cite{lee2018deephit}, Neural Multi-Task Logistic Regression (N-MTLR)~\cite{fotso2018deep}, Nnet-Survival~\cite{gensheimer2019scalable}, and Piecewise-Constant Hazard (PC-Hazard)~\cite{kvamme2021continuous}.
Hyperparameters for FedSurF and FedSurF-IBS are optimized using cross-validation.
For CoxPH, the number of parameters is equal to the number of features in the input space.
The neural network architectures of DeepSurv, DeepHit, N-MTLR, Nnet-Survival, and PC-Hazard are composed of two dense layers with 32 neurons each, followed by a ReLU activation function and a Dropout layer to prevent overfitting. 
The last dense layer of DeepSurv produces a scalar output compliant with the proportional hazard assumption while the other models had 10 outputs, one for each of the discretization instants.

\subsection{Training}

In the experiments, we trained each model both in a local setting and in a federated setting.
In the local setting, clients train and validate their models on local data only.
Local evaluation is useful to understand whether joining a federated learning algorithm is beneficial for a client.

The implementation of RSF is based on the scikit-survival Python library~\cite{sksurv}, while the other differentiable models rely on PyCox~\cite{kvamme2019time}.
RSF hyperparameter tuning is conducted with cross-validation.
Differentiable models are trained using the Adam optimizer with a learning rate of 0.01 and early stopping for 500 epochs at most. 
For FedSurF and FedSurF-IBS, federated training is performed according to the algorithm outlined in Section~\ref{sec:federated-survival-forests}.
Conversely, FedAvg~\cite{mcmahan2017communication} is applied to all the other models. 
Concerning proportional hazard models (CoxPH and DeepSurv), we assume each client has access to the same global Nelson-Aalen estimation of the baseline hazard.
To address the issue of convergence in heterogeneous federations, we ran FedAvg for 500 rounds with a single local epoch per round and selected the parameters with the best validation metrics for the final evaluation on the test set.

\subsection{Evaluation}
\label{sec:eval}

In our evaluation, we utilized the Concordance Index (C-Index) and the Integrated Brier Score (IBS) to assess the performance of our models on the test set. 
The C-Index~\cite{uno2011c} measures the proportion of comparable samples for which the prediction and true outcome are in agreement. 
A sample pair is considered comparable if at least one of the two samples is not censored. 
The C-Index reflects the discriminatory power of the model, indicating the fraction of samples that the model has ordered correctly according to actual time outcomes. 
A random guessing model would have a C-Index of 0.5, while a model with perfect knowledge would have a C-Index of 1.0.

The Brier score~\cite{graf1999assessment} is a calibration metric that measures the squared difference between the true survival status and the predicted survival probability of the model at a given time. 
The IBS integrates the Brier score over time to condense the calibration level of a model into a single value. 
A random guessing model would have an IBS of 0.25, while a model with perfect knowledge would have an IBS of 0.0. Thus, the lower the IBS, the better the model.

To account for the censoring distribution, we applied Inverse Probability of Censoring Weighting (IPCW)~\cite{robins1992recovery,uno2011c} during metrics estimation to ensure unbiased results.

\subsection{Results}
\begin{table*}[ht!]
\centering
\caption{C-Index for survival models evaluated on uniformly split datasets. Each metric is scaled by a factor of 100 for readability. We report the mean and standard deviation over 5 runs. The best results ($\uparrow$) are highlighted in bold.}
\label{tab:uniform-concordance}
    
\begin{tabularx}{\textwidth}{@{} l YY YY YY YY YY @{}}
	\toprule
	Dataset & \multicolumn{2}{c}{GBSG2} & \multicolumn{2}{c}{METABRIC} & \multicolumn{2}{c}{AIDS} & \multicolumn{2}{c}{FLCHAIN} & \multicolumn{2}{c}{SUPPORT}\\
	\cmidrule(r){1-1}\cmidrule(l){2-3}\cmidrule(l){4-5}\cmidrule(l){6-7}\cmidrule(l){8-9}\cmidrule(l){10-11}
	Model & \emph{Local} & \emph{Federated} & \emph{Local} & \emph{Federated} & \emph{Local} & \emph{Federated} & \emph{Local} & \emph{Federated} & \emph{Local} & \emph{Federated}\\ 
	\midrule
	\mbox{CoxPH} & 61.2$\pm$4.1 & 68.7$\pm$8.5 & 58.8$\pm$1.9 & 62.0$\pm$1.7 & 51.9$\pm$0.7 & 53.8$\pm$1.6 & 92.3$\pm$1.6 & 93.4$\pm$0.4 & 79.9$\pm$0.2 & 82.5$\pm$0.3 \\ 
	\mbox{DeepSurv} & 60.7$\pm$4.0 & 69.3$\pm$7.5 & 59.7$\pm$1.7 & \textbf{63.0$\pm$2.7} & 52.5$\pm$0.2 & 53.7$\pm$1.3 & 93.5$\pm$0.4 & 94.2$\pm$0.2 & 79.9$\pm$0.4 & 83.1$\pm$0.4 \\ 
	\mbox{DeepHit} & 59.8$\pm$5.3 & 68.5$\pm$7.5 & 57.3$\pm$1.3 & 61.1$\pm$3.0 & 52.6$\pm$0.5 & 53.8$\pm$1.6 & 93.1$\pm$0.3 & 93.8$\pm$0.4 & 81.7$\pm$0.4 & 83.5$\pm$0.6 \\ 
	\mbox{N-MTLR} & 61.5$\pm$4.4 & 69.2$\pm$7.1 & 58.3$\pm$1.6 & 62.2$\pm$2.5 & 53.1$\pm$0.8 & \textbf{54.2$\pm$1.6} & 93.6$\pm$0.3 & \textbf{94.2$\pm$0.3} & 80.6$\pm$0.4 & 83.0$\pm$0.6 \\ 
	\mbox{Nnet-Survival} & 60.5$\pm$7.9 & 66.7$\pm$7.7 & 49.4$\pm$3.9 & 60.4$\pm$4.1 & 50.8$\pm$1.2 & 54.0$\pm$1.6 & 90.7$\pm$1.3 & 94.1$\pm$0.4 & 80.9$\pm$0.3 & \textbf{83.8$\pm$0.6} \\ 
	\mbox{PC-Hazard} & 60.3$\pm$8.2 & 67.3$\pm$7.4 & 50.2$\pm$3.9 & 59.8$\pm$3.8 & 51.0$\pm$1.3 & 53.3$\pm$1.8 & 89.8$\pm$0.8 & 94.1$\pm$0.3 & 80.7$\pm$0.3 & 81.9$\pm$1.2 \\ 
	\midrule
	\mbox{FedSurF} & 66.5$\pm$5.7 & \textbf{72.3$\pm$4.7} & 59.9$\pm$2.1 & 62.2$\pm$1.9 & 52.4$\pm$0.6 & 54.1$\pm$1.3 & 93.4$\pm$0.3 & 93.5$\pm$0.3 & 80.4$\pm$0.4 & 81.0$\pm$0.4 \\ 
	\mbox{FedSurF-IBS} & -- & 71.8$\pm$5.4 & -- & 62.4$\pm$2.0 & -- & 53.8$\pm$1.3 & -- & 93.5$\pm$0.3 & -- & 80.7$\pm$0.6 \\ 
	\bottomrule
\end{tabularx}

\end{table*}

\begin{table*}[ht!]
\centering
\caption{IBS for survival models evaluated on uniformly split datasets. Each metric is scaled by a factor of 100 for readability. We report the mean and standard deviation over 5 runs. The best results ($\downarrow$) are highlighted in bold.}
\label{tab:uniform-ibs}
    
\begin{tabularx}{\textwidth}{@{} l YY YY YY YY YY @{}}
	\toprule
	Dataset & \multicolumn{2}{c}{GBSG2} & \multicolumn{2}{c}{METABRIC} & \multicolumn{2}{c}{AIDS} & \multicolumn{2}{c}{FLCHAIN} & \multicolumn{2}{c}{SUPPORT}\\
	\cmidrule(r){1-1}\cmidrule(l){2-3}\cmidrule(l){4-5}\cmidrule(l){6-7}\cmidrule(l){8-9}\cmidrule(l){10-11}
	Model & \emph{Local} & \emph{Federated} & \emph{Local} & \emph{Federated} & \emph{Local} & \emph{Federated} & \emph{Local} & \emph{Federated} & \emph{Local} & \emph{Federated}\\ 
	\midrule
	\mbox{CoxPH} & 20.2$\pm$1.1 & 18.5$\pm$1.6 & 17.9$\pm$1.9 & 17.4$\pm$2.3 & 15.1$\pm$1.5 & 14.7$\pm$1.5 & 5.6$\pm$0.4 & 5.4$\pm$0.2 & 13.3$\pm$0.2 & 12.7$\pm$0.1 \\ 
	\mbox{DeepSurv} & 21.2$\pm$1.2 & \textbf{17.9$\pm$1.2} & 17.8$\pm$1.5 & \textbf{16.6$\pm$1.5} & 15.2$\pm$1.2 & 14.7$\pm$1.5 & 5.6$\pm$0.6 & 4.2$\pm$0.1 & 13.9$\pm$0.2 & 12.2$\pm$0.2 \\ 
	\mbox{DeepHit} & 21.8$\pm$2.0 & 19.5$\pm$1.1 & 18.7$\pm$1.7 & 17.6$\pm$1.6 & 16.0$\pm$1.4 & 14.8$\pm$1.5 & 5.8$\pm$0.2 & 4.5$\pm$0.1 & 13.7$\pm$0.6 & 12.7$\pm$0.4 \\ 
	\mbox{N-MTLR} & 21.3$\pm$1.6 & 18.3$\pm$1.0 & 18.1$\pm$1.5 & 17.1$\pm$1.7 & 14.9$\pm$1.4 & 14.7$\pm$1.4 & 4.8$\pm$0.1 & \textbf{4.2$\pm$0.1} & 14.5$\pm$0.3 & \textbf{11.9$\pm$0.4} \\ 
	\mbox{Nnet-Survival} & 23.0$\pm$2.1 & 19.5$\pm$1.0 & 23.9$\pm$2.7 & 17.7$\pm$1.6 & 15.4$\pm$1.5 & 15.0$\pm$1.5 & 6.5$\pm$0.4 & 4.5$\pm$0.2 & 12.9$\pm$0.3 & 12.3$\pm$0.4 \\ 
	\mbox{PC-Hazard} & 22.4$\pm$1.9 & 19.6$\pm$0.8 & 23.7$\pm$2.4 & 18.0$\pm$1.6 & 15.4$\pm$1.6 & 15.1$\pm$1.5 & 6.8$\pm$0.4 & 4.5$\pm$0.2 & 13.0$\pm$0.3 & 12.4$\pm$0.4 \\ 
	\midrule
	\mbox{FedSurF} & 19.3$\pm$1.2 & 18.5$\pm$1.2 & 17.6$\pm$1.3 & 16.8$\pm$1.2 & 14.8$\pm$1.5 & 14.7$\pm$1.5 & 4.6$\pm$0.1 & 4.5$\pm$0.1 & 18.2$\pm$0.5 & 18.1$\pm$0.5 \\ 
	\mbox{FedSurF-IBS} & -- & 18.4$\pm$1.2 & -- & 16.7$\pm$1.3 & -- & \textbf{14.7$\pm$1.5} & -- & 4.4$\pm$0.1 & -- & 15.9$\pm$0.5 \\ 
	\bottomrule
\end{tabularx}

\end{table*}

\begin{table*}[ht!]
\centering
\caption{C-Index for survival models evaluated on label-skewed datasets. Each metric is scaled by a factor of 100 for readability. We report the mean and standard deviation over 5 runs. The best results ($\uparrow$) are highlighted in bold.}
\label{tab:skew-concordance}
    
\begin{tabularx}{\textwidth}{@{} l YY YY YY YY YY @{}}
	\toprule
	Dataset & \multicolumn{2}{c}{GBSG2} & \multicolumn{2}{c}{METABRIC} & \multicolumn{2}{c}{AIDS} & \multicolumn{2}{c}{FLCHAIN} & \multicolumn{2}{c}{SUPPORT}\\
	\cmidrule(r){1-1}\cmidrule(l){2-3}\cmidrule(l){4-5}\cmidrule(l){6-7}\cmidrule(l){8-9}\cmidrule(l){10-11}
	Model & \emph{Local} & \emph{Federated} & \emph{Local} & \emph{Federated} & \emph{Local} & \emph{Federated} & \emph{Local} & \emph{Federated} & \emph{Local} & \emph{Federated}\\ 
	\midrule
	\mbox{CoxPH} & 60.3$\pm$2.8 & 68.4$\pm$7.8 & 57.3$\pm$0.9 & 60.6$\pm$3.8 & 52.2$\pm$1.4 & 54.2$\pm$1.1 & 91.5$\pm$2.8 & 77.6$\pm$22.5 & 80.0$\pm$0.2 & 82.3$\pm$0.3 \\ 
	\mbox{DeepSurv} & 59.8$\pm$2.2 & 67.5$\pm$8.5 & 58.7$\pm$1.8 & 62.1$\pm$3.7 & 53.0$\pm$0.6 & 54.5$\pm$1.6 & 92.3$\pm$3.1 & 81.9$\pm$18.6 & 80.0$\pm$0.5 & 80.7$\pm$2.2 \\ 
	\mbox{DeepHit} & 60.9$\pm$5.3 & 68.5$\pm$7.0 & 56.2$\pm$1.0 & 60.4$\pm$2.9 & 53.2$\pm$1.1 & 53.9$\pm$0.9 & 93.2$\pm$0.3 & 93.7$\pm$0.1 & 81.0$\pm$0.2 & 82.9$\pm$0.6 \\ 
	\mbox{N-MTLR} & 61.8$\pm$8.9 & 69.3$\pm$7.6 & 59.0$\pm$1.9 & 61.7$\pm$3.3 & 53.0$\pm$0.7 & 53.6$\pm$2.1 & 93.6$\pm$0.2 & \textbf{94.0$\pm$0.2} & 80.3$\pm$0.3 & 81.1$\pm$0.4 \\ 
	\mbox{Nnet-Survival} & 59.5$\pm$8.5 & 68.2$\pm$7.7 & 49.1$\pm$3.6 & 59.5$\pm$3.7 & 51.3$\pm$1.5 & 54.2$\pm$2.3 & 90.6$\pm$1.2 & 93.9$\pm$0.3 & 80.7$\pm$0.5 & \textbf{83.5$\pm$0.7} \\ 
	\mbox{PC-Hazard} & 60.9$\pm$8.0 & 68.9$\pm$7.3 & 49.4$\pm$3.8 & 59.2$\pm$3.6 & 50.9$\pm$0.6 & 53.6$\pm$2.6 & 89.9$\pm$1.1 & 93.9$\pm$0.3 & 80.6$\pm$0.5 & 80.6$\pm$0.8 \\ 
	\midrule
	\mbox{FedSurF} & 66.8$\pm$5.4 & 71.9$\pm$5.0 & 59.9$\pm$1.4 & 61.8$\pm$2.9 & 53.2$\pm$0.8 & 55.0$\pm$0.9 & 93.4$\pm$0.3 & 93.5$\pm$0.3 & 80.3$\pm$0.4 & 81.3$\pm$0.4 \\ 
	\mbox{FedSurF-IBS} & -- & \textbf{72.4$\pm$5.4} & -- & \textbf{62.1$\pm$2.4} & -- & \textbf{55.1$\pm$0.6} & -- & 93.5$\pm$0.2 & -- & 80.6$\pm$0.7 \\ 
	\bottomrule
\end{tabularx}

\end{table*}

\begin{table*}[ht!]
\centering
\caption{IBS for survival models evaluated on label-skewed datasets. Each metric is scaled by a factor of 100 for readability. We report the mean and standard deviation over 5 runs. The best results ($\downarrow$) are highlighted in bold.}
\label{tab:skew-ibs}
    
\begin{tabularx}{\textwidth}{@{} l YY YY YY YY YY @{}}
	\toprule
	Dataset & \multicolumn{2}{c}{GBSG2} & \multicolumn{2}{c}{METABRIC} & \multicolumn{2}{c}{AIDS} & \multicolumn{2}{c}{FLCHAIN} & \multicolumn{2}{c}{SUPPORT}\\
	\cmidrule(r){1-1}\cmidrule(l){2-3}\cmidrule(l){4-5}\cmidrule(l){6-7}\cmidrule(l){8-9}\cmidrule(l){10-11}
	Model & \emph{Local} & \emph{Federated} & \emph{Local} & \emph{Federated} & \emph{Local} & \emph{Federated} & \emph{Local} & \emph{Federated} & \emph{Local} & \emph{Federated}\\ 
	\midrule
	\mbox{CoxPH} & 20.1$\pm$1.2 & 18.5$\pm$1.4 & 18.3$\pm$1.6 & 17.3$\pm$2.3 & 15.1$\pm$1.5 & 14.7$\pm$1.5 & 5.7$\pm$0.6 & 8.5$\pm$4.0 & 13.5$\pm$0.2 & 12.8$\pm$0.2 \\ 
	\mbox{DeepSurv} & 20.6$\pm$1.5 & \textbf{18.2$\pm$1.0} & 18.7$\pm$1.6 & \textbf{16.8$\pm$1.7} & 15.2$\pm$1.1 & 14.7$\pm$1.5 & 5.5$\pm$0.6 & 9.0$\pm$4.4 & 14.0$\pm$0.3 & 12.6$\pm$0.5 \\ 
	\mbox{DeepHit} & 21.7$\pm$2.0 & 19.2$\pm$0.7 & 18.6$\pm$1.7 & 17.7$\pm$1.5 & 16.0$\pm$1.3 & 14.7$\pm$1.5 & 5.7$\pm$0.2 & 4.5$\pm$0.1 & 14.4$\pm$0.6 & 12.8$\pm$0.4 \\ 
	\mbox{N-MTLR} & 20.8$\pm$0.9 & 18.7$\pm$1.5 & 18.2$\pm$1.7 & 17.4$\pm$1.9 & 15.0$\pm$1.4 & 14.7$\pm$1.4 & 4.8$\pm$0.1 & 4.5$\pm$0.2 & 14.7$\pm$0.3 & 13.6$\pm$0.7 \\ 
	\mbox{Nnet-Survival} & 23.6$\pm$2.5 & 19.1$\pm$1.5 & 23.9$\pm$2.4 & 17.9$\pm$1.4 & 15.4$\pm$1.6 & 14.9$\pm$1.6 & 6.6$\pm$0.5 & 4.5$\pm$0.2 & 13.0$\pm$0.4 & \textbf{12.6$\pm$0.6} \\ 
	\mbox{PC-Hazard} & 22.6$\pm$2.1 & 19.0$\pm$1.5 & 23.9$\pm$2.5 & 18.2$\pm$1.6 & 15.4$\pm$1.5 & 15.0$\pm$1.6 & 6.8$\pm$0.4 & 4.5$\pm$0.2 & 13.1$\pm$0.4 & 12.7$\pm$0.5 \\ 
	\midrule
	\mbox{FedSurF} & 19.3$\pm$1.2 & 18.5$\pm$1.3 & 17.8$\pm$1.2 & 16.9$\pm$1.3 & 14.9$\pm$1.5 & 14.7$\pm$1.5 & 4.7$\pm$0.1 & 4.5$\pm$0.1 & 18.5$\pm$0.6 & 18.0$\pm$0.6 \\ 
	\mbox{FedSurF-IBS} & -- & 18.4$\pm$1.1 & -- & 16.9$\pm$1.2 & -- & \textbf{14.7$\pm$1.5} & -- & \textbf{4.4$\pm$0.1} & -- & 15.8$\pm$0.5 \\ 
	\bottomrule
\end{tabularx}

\end{table*}

The performance of the compared survival models is summarized in Table~\ref{tab:uniform-concordance} and Table~\ref{tab:uniform-ibs} for federations built with uniform data splitting, and in Table~\ref{tab:skew-concordance} and Table~\ref{tab:skew-ibs} for federations built with label-skewed splitting. 
In the following, we provide a detailed analysis of the results obtained.

\emph{Is it beneficial for a client to join a federated learning procedure?}
All the results agree that on average the best performance is consistently achieved under the \emph{Federated} columns. 
This indicates that, for a given survival model, it is generally advantageous to join a federation in terms of both discrimination and calibration.

\emph{Is FedSurF performing better than the alternatives when data is uniformly distributed?} 
To answer this question, we refer to Table~\ref{tab:uniform-concordance} and Table~\ref{tab:uniform-ibs}. 
From the results, it can be seen that FedSurF does not consistently outperform other state-of-the-art models. However, its performance is comparable to that of the other models in terms of both discrimination and calibration.
Upon closer examination, we observe that FedSurF performs particularly well for smaller datasets (GBSG2 and METABRIC), while for larger datasets (AIDS, FLCHAIN, and SUPPORT), the results are comparable to those of deep learning models. 
This suggests that the model choice between ensemble-based and neural-based is not particularly relevant when data is uniformly distributed. 
However, we highlight that FedSurF and FedSurF-IBS obtain the recorded performance in just a single communication round, which may be a relevant advantage in federations with communication channel constraints.

\emph{Is IBS sampling relevant when data is uniformly distributed?} 
When data is uniformly distributed FedSurF-IBS does not consistently outperform FedSurF. 
This can be seen by comparing the results in Table~\ref{tab:uniform-concordance} and Table~\ref{tab:uniform-ibs}. 
This finding suggests that the use of IBS sampling is not necessary for achieving optimal performance in scenarios with identically distributed data.
This outcome is expected, as IBS sampling is particularly useful to select the best trees from the best clients. 
In the case of IID data, where the local data distribution is roughly the same, the quality of the trees generated by each client should be similar. 
Under these conditions, IBS does not provide an advantage for achieving optimal performance.

\emph{Is FedSurF performing better than the alternatives when data is heterogeneous?}
To comment on this, we refer to Table~\ref{tab:skew-concordance} and Table~\ref{tab:skew-ibs}.
The results show that FedSurF-IBS obtains the best or very close to the best metrics on all datasets except for SUPPORT. 
Specifically, it is the model with the best C-Index on GBSG2, METABRIC, and AIDS. 
Additionally, the IBS is also very low and within the standard deviation of the slightly better models.
From these results, it can be concluded that FedSurF with IBS sampling is a well-performing algorithm in heterogeneous federations when considering the C-Index and IBS metrics. 
It demonstrates a robust performance across different label-skewed datasets, with a consistent improvement over the other models.

\emph{Is IBS sampling relevant when data is heterogeneous?} 
The results for the heterogeneous datasets, as presented in Table~\ref{tab:skew-concordance} and Table~\ref{tab:skew-ibs}, show that for all datasets except for SUPPORT, FedSurF-IBS outperforms FedSurF.
This indicates that, in contrast to federations with uniformly split data, IBS sampling is relevant and beneficial in federations where data is heterogeneous. 
Indeed, it allows to select the best trees from the best clients and improve the overall performance of the ensemble model. 
This highlights the importance of considering the characteristics of the data distribution when choosing the appropriate sampling method for FedSurF.
Concerning the results of the SUPPORT dataset, despite an extensive hyperparameter search, survival forests exhibit a higher Brier Score than neural models in 10-client federations.
We impute the lower IBS to the tree-based nature of the models, which may model the specific survival rates of SUPPORT suboptimally with respect to neural-based methods.
This claim is confirmed by the \emph{Local} results, which highlight how, on average, tree-based models have lower performance than neural models, even without considering federated learning.
Nevertheless, applying IBS sampling in the \emph{Federated} experiments lowers the IBS by more than 0.02, yielding a significant improvement over uniform sampling.

\section{Conclusion}
\label{sec:conclusion}

This work presents Federated Survival Forest (FedSurF), an extension of the Random Survival Forest (RSF) algorithm to the federated learning setting. 
Our proposed algorithm is based on sampling the best local trees from the best clients in the federation to build a RSF on the server.
We demonstrate the effectiveness of FedSurF through extensive experiments in two different federated settings: one with uniformly split data and another with heterogeneous data. 
Our results indicate that FedSurF performs comparably well to state-of-the-art models when data is uniformly distributed while obtaining a noticeable advantage in heterogeneous federations. 
Furthermore, FedSurF relies on a single communication round between clients and server, rather than multiple iterative updates, making it an efficient and practical solution in federations with strict bandwidth requirements.

\section*{Acknowledgment}
This project has been supported by AI-SPRINT: AI in Secure Privacy-pReserving computINg conTinuum (European Union's~H2020 grant agreement No. 101016577) and FAIR: Future Artificial Intelligence Research (NextGenerationEU, PNRR-PE-AI scheme, M4C2, investment 1.3, line on Artificial Intelligence).

\bibliographystyle{IEEEtran}
\bibliography{IEEEabrv,references}

\end{document}